\documentclass[review]{elsarticle}

\usepackage{hyperref}
\usepackage{amsmath,graphicx}
\usepackage{booktabs}
\usepackage{url}  
\usepackage{rotating}
\usepackage{color,soul}
\usepackage{subfig}
\usepackage{float}
\usepackage{ifoddpage}
\usepackage{tcolorbox}
\AtBeginEnvironment{tcolorbox}{\small}
\usepackage{adjustbox}
\usepackage{mathtools}
\usepackage{pgfplots}
\usepackage{eqparbox}
\usepackage{adjustbox}
\usepackage{tikz}
\usepackage{amsfonts}
\usepackage{pgfplotstable}
\usetikzlibrary{patterns}
\usetikzlibrary{matrix}
\usepgfplotslibrary{groupplots}
\pgfplotsset{compat=newest}
\usepackage{soul}
\usepackage{color}
\def\x{{\mathbf x}}

\usepackage{verbatim}
\usepackage{smartdiagram}
\pgfplotsset{every tick label/.append style={font=\tiny}}
 \pgfplotsset{
    axis line style={black, very thick, -stealth},
    every axis label/.append style ={black},
    every tick label/.append style={black}  
  }

\journal{Artificial Intelligence Journal}

\usetikzlibrary{arrows.meta}
\tikzset{>={Latex[round]}}
\begin{document}

\begin{frontmatter}

\title{A multi-component framework for the analysis and design of explainable artificial intelligence}
\author{S. Atakishiyev,  H. Babiker,  N. Farruque,  R. Goebel\footnote{Authors listed in alphabetical order; R. Goebel (rgoebel@ualberta.ca) is the corresponding author.}, M-Y. Kim$^{a}$, M.H. Motallebi,  J. Rabelo,  T. Syed, O. R. Za\"{\i}ane}
\address{XAI Lab, \\
Alberta Machine Intelligence Institute,\\
Department of Computing Science,\\
University of Alberta, \\
Edmonton, Alberta, Canada T6G 2E8}
\address[a]{Department of Science, \\
Augustana Faculty, University of Alberta,\\
Camrose, Alberta, Canada, T4V 2R3}

\begin{abstract}
The rapid growth of research in explainable artificial intelligence (XAI) follows on two substantial developments. First, the enormous application success of modern machine learning methods, especially deep and reinforcement learning, which have created high expectations for industrial, commercial and social value.  Second, the emergence of concern for creating trusted AI systems, including the creation of regulatory principles to ensure transparency and trust of AI systems.

These two threads have created a kind of ``perfect storm'' of research activity, all eager to create and deliver {\it any} set of tools and techniques to address the XAI demand.

As some surveys of current XAI suggest, there is yet to appear a principled framework that respects the literature of explainability in the history of science, and which provides a basis for the development of a framework for transparent XAI.  Here we intend to provide a strategic inventory of XAI requirements, demonstrate their connection to a history of XAI ideas, and synthesize those ideas into a simple framework to calibrate five successive levels of XAI.

\end{abstract}


\end{frontmatter}


\section{Introduction}
\label{section:intro}

Fueled by a growing need for trust and ethical artificial intelligence (AI) by design, the wake of the last decade of machine learning is crowded with a broad spectrum of research on explainable AI (XAI). 

It is therefore not surprising that the rapid and eclectic flurry of activities in XAI have exposed confusion and controversy about foundational concepts like {\it interpretability}, {\it explanation}, and {\it causality}.

Perhaps confusion and disagreement is not surprising, given some of the complexity of modern learning methods.  For example, when some deep learning methods can build predictive models based on more than 50 million distinct parameters, it is not a surprise that humans will debate what has been captured (e.g., see \cite{Chollet2017}\footnote{Especially see Section 2, Chapter 9, {\it The limitations of deep learning.}}). Note also the confusion regarding misconceptions on a specious trade off between predictive accuracy and explainability (cf. \cite{rudin2018stopexplaining}), which have precipitated scientific workshops to address such misconceptions\footnote{Note the 32nd Conference on Neural Information Processing Systems (NeurIPS 2018), Workshop on Critiquing and Correcting Trends in Machine Learning, https://ml-critique-correct.github.io}. Another example of yet-to-be-resolved issues includes the strange anomaly where syntactic reduction of the parameter space of some deep learning created models actually results in improved predictive accuracy (e.g., \cite{molchanov2019pruning,cohen2018overparamterization}). The reality is that the foundations for scientific understanding of general machine learning, and thus XAI, is not yet sufficiently developed.

Even though the long history of science and more recent history of scientific explanation and causality have considerable contributions to make (e.g., \cite{woodward2003scientific,pearl2018book}), it seems like the demand created by potential industrial value has induced brittleness in identifying and confirming a robust trajectory from the history of formal systems to modern applications of AI.  However, we believe that one can re-establish some important scientific momentum by exploiting what Newell and Simon's Turing Award paper (\cite{newellsimon1976}) identified as the physical symbol systems hypothesis: ``A physical symbol system has the necessary and sufficient means for general intelligent action." (p. 116).

The challenge is to clarify connections between the recent vocabulary of XAI and their historical roots, in order to distinguish between scientifically valuable history and potentially new extensions.  In what follows, we hope to articulate and connect a broader historical fabric of concepts essential to XAI, including interpretation, explanation,  causality, evaluation, system debugging, expressivity, semantics, inference, abstraction, and prediction.

\pgfplotstableread{
Attributes     Explicit-explanation-representation Alternative-explanations Knowledge-of-the-explainee Interactive
 Level-0     0        0        0       0
 Level-1     30       0        0       0
 Level-2     30       30       0       0
 Level-3     30       30      30       0
 Level-4     30       30      30       30
}{\data}

\pgfplotstablecreatecol[
    create col/expr={
        \thisrow{Explicit-explanation-representation} + \thisrow{Alternative-explanations} + \thisrow{Knowledge-of-the-explainee} + \thisrow{Interactive}
    }
]{Sum}{\data}

\pgfplotstablesave[columns/Attributes/.style={string type}, columns/Sum/.style={numeric as string type}, col sep=comma, disable rowcol styles=false]{\data}{temptable.txt}

\makeatletter
\begin{figure*}
\centering
\begin{tikzpicture}[scale=0.9]
  \begin{axis}[
  scale only axis,
axis line style = thick,
         axis x line=bottom,
         axis y line=left,
         inner axis line style={>={Latex[round]},very thick,black},
         axis line style={-Latex[round]},
every axis x label/.style={
    at={(ticklabel* cs:1.05)},
    anchor=west,
},
every axis y label/.style={
    at={(ticklabel* cs:1.05)},
    anchor=south,
},
legend cell align={left},
yticklabels={,,},
    ymin=0,
    width=0.7\textwidth,
    enlarge x limits=0.30,
    symbolic x coords={Level-0, Level-1, Level-2, Level-3, Level-4},
    xtick=data,
    bar width=7mm,
        ylabel={Attributes },
    xlabel={ Scale of explanation },
    point meta=explicit,
    calculate offset/.code={
        \pgfkeys{/pgf/fpu=true,/pgf/fpu/output format=fixed}
        \pgfmathsetmacro\testmacro{(\pgfplotspointmetatransformed)/1000)}
        \pgfkeys{/pgf/fpu=false}
    },
    every node near coord/.style={
        /pgfplots/calculate offset,
        yshift=-\testmacro,
        xshift=-5pt,
        font=\tiny
    },
    legend pos = north west,
    legend style={
      draw=none },
    nodes near coords align=center,
    ybar stacked, 
        legend style={font=\fontsize{3}{4}\selectfont},
    ]

    \addplot+[line width=1pt,solid,    black,
    fill=yellow,
    postaction={
        pattern=north east lines
    }
    ] table[y=Explicit-explanation-representation, meta=Explicit-explanation-representation, col sep=comma] {temptable.txt};
    \addplot+[line width=1pt,solid,    black,
    fill=lime,
    postaction={
        pattern=grid
    }
    ] table[y=Alternative-explanations, meta=Alternative-explanations, col sep=comma] {temptable.txt};
    \addplot+[line width=1pt,solid,    black,
    fill=cyan,
    postaction={
        pattern=vertical lines
    }
    ] table[y=Knowledge-of-the-explainee, meta=Knowledge-of-the-explainee, col sep=comma] {temptable.txt};
        \addplot+[line width=1pt,solid,    black,
    fill=gray,
    postaction={
        pattern=dots
    }
    ] table[y=Interactive, meta=Interactive, col sep=comma] {temptable.txt};
\legend{ Explicit-explanation-representation,    Alternative-explanations, Knowledge-of-the-explainee, Interactive }
  \end{axis}
\end{tikzpicture}
\caption{Major Explanatory Components and their Potential Role in a Scale of Explanation}
\label{fig:explainability-diagram}
\end{figure*}
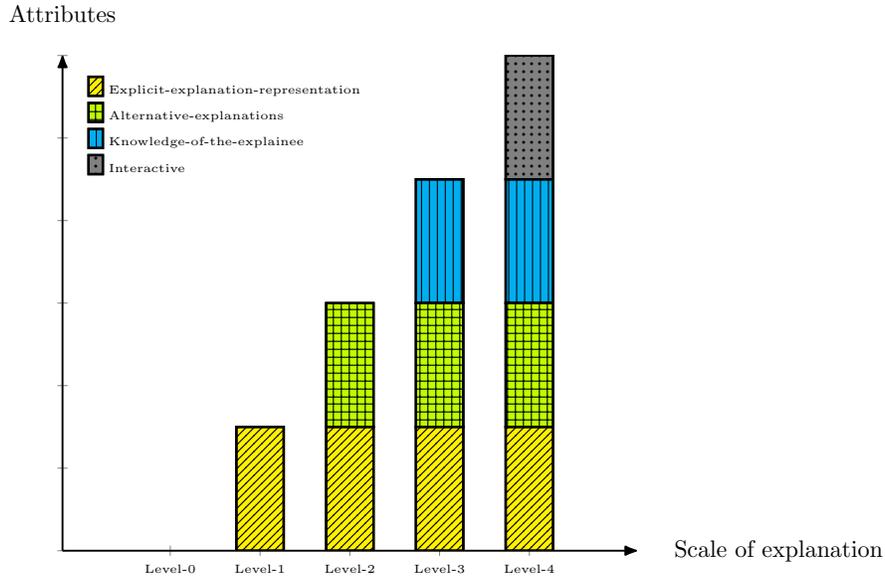
\makeatother

\begin{table}
    \begin{center}
       \resizebox{0.9\textwidth}{!}{%
            \begin{tabular}{l|ccccc}
                Attributes                              & Level 0 & Level 1 & Level 2 & Level 3 & Level 4 \\
                \hline
                Explicit Explanation Representation     &  &\checkmark  &\checkmark  &\checkmark  &\checkmark   \\
                Alternative Explanations                &  &            &\checkmark  &\checkmark  &\checkmark   \\
                Knowledge of the explainee              &  &            &            &\checkmark  &\checkmark   \\
                Interactive                             &  &            &            &            &\checkmark   \\
           \end{tabular}
        }
    \end{center}
    \caption{Tabular Representation of Levels of Explanation and the Corresponding Attributes}
    \label{tab:tabular_level_explainability}
\end{table}

To do so, we need to articulate a general, if incomplete XAI framework, which is immediately challenged to be comprehensive enough to acknowledge a broad spectrum of components, yet but still avoid becoming a na\"{\i}ve long and unstructured survey of overlapping -- sometimes competing -- concepts and methods.  Our approach is to start with a simple structural idea first illustrated by Figure~\ref{fig:explainability-diagram}.

One can conceptualize the content of Figure~\ref{fig:explainability-diagram} by comparing an accepted framework around the articulation levels of autonomous driving (\cite{SAE2016}). The simple distinction of degree of human control creates the basis for discussion about how one could achieve Level~0 (no automation) to Level~5 (full automation), then discuss the details of those levels and the components necessary to achieve them.

In our case for explainability, we intend the x-axis of Figure~\ref{fig:explainability-diagram}  to distinguish what we consider a kind of quality of explanatory system, with the intuition that explanations at a higher level can be confirmed as ``better'' by some evaluation measure, e.g., we expect a causal explanation to provide the basis for recreating a causal inference chain to a prediction.  We would distinguish, for example ``explanation by authority,'' towards the left end of the scale, to be something like an explanation to the question ``why can't I go out tonight?'' to be something like ``Because I said so,'' from a parent to a teenager, which we might just say is explanation by authority.  Just these simple distinctions help frame and motivate a more detailed analysis of distinctions across our informal scale of levels of explanation, which will be further articulated in what follows.

Note that an alternative to Figure~\ref{fig:explainability-diagram}, might be a simple abstract comparison of levels of explanation as a check list of possible components, like in Table~\ref{tab:tabular_level_explainability}. Please note that both representations are intended only to begin to consider how such components may obtain in any particular XAI system.  For example, our figure and table do not intend the reader to draw the inference that an XAI system is somehow better with a dialogue component than without, nor that any anticipated evaluation of performance is higher with an integration of more of the components.   The idea is only to suggest that there will emerge a foundation of what XAI components are essential, orthogonal, and have distinct and value-contributing roles in future XAI systems.

Similarly, our desire for some kind of sensible measures to clarify a level $n$ explanation from a level $n+1$ explanation creates our speculation on a kind of y-axis, which is not intended to imply the existence of any measure.  Rather our y-axis is a kind of independent set of plausible orthogonal explanatory system attributes, which should be distinguished clearly enough to be able to use each attribute as a check list of attributes that any explanatory system may or may not have (cf. Table~\ref{tab:tabular_level_explainability}).

For example, most surveys of XAI note the requirement for a system to produce alternative explanations; simply put, producing a single explanation may be completely insufficient for multiple explainees \cite{miller-explanation-2019}.  Similarly, many have noted the value of interactive XAI systems and dialogue systems \cite{Bex2016combining}, which provide a basis for an explainee to submit and receive responses to questions about a model prediction, and thus build deeper trust of the system.

In what follows, we will provide increasing detail and precision about how we believe existing XAI concepts align with this simple framework, in order to consider how to articulate choices in the design of an XAI system.

The rest of our paper is organized as follows.  Section~\ref{section:principlecomponents} 
presents what we consider the principal components of XAI, including that explanations need to be explainee-specific, that there can always be multiple explanations, and that the evaluation of the quality of explanation has more to do with the explainee than the explanation system. This will help provide sufficient detail to articulate the relationship between current explanatory concepts and their relationship to historical roots, e.g., to consider the emerging demands on the properties of a formal definition of {\it interpretability} by assessing the classical formal systems view of interpretability.  Section~\ref{section:history} considers the more general history of explanation, as an attempt to connect the formal philosophy and scientific explanation foundations. This concludes with the articulation of the explanatory role of recent theories of causal representation.  Section~\ref{section:trends} summarizes important emerging trends and components in proposed XAI architectures, including those that apply to both machine-learned predictive models and general AI systems.   Section~\ref{section:components} provides a synopsis of current XAI research threads, and how they might integrate into an emerging XAI architecture.  The 
opinions include the description of important XAI ideas like pre-hoc versus post-hoc explanation creation, and the evaluation of explanations, in order to sketch an architecture of how necessary components for XAI are connected. Finally Section~\ref{section:conclusions} provides a brief summary, and what we believe the future architectures of XAI systems will require to ensure the trust of future AI systems.


\section{Principal Components at the foundations of XAI}
\label{section:principlecomponents}

\subsection{Explainability and Interpretability}

There is a confusion in the literature regarding the definitions of interpretability and explainability of models. Many recent papers use those terms interchangeably \cite{koh2017understanding}, \cite{vaughan2018explainable}. Some papers do make a distinction between those terms, but we do not agree with those definitions as well. For example, Gilpin et al. \cite{gilpin2018explaining} define interpretability as a rudimentary form of explainability. Rudin \cite{rudin2019stop} finds that there is no single definition on interpretability. However, the author defines a spectrum which extends from fully interpretable models such as rule-based models (that provide explanations by definition) to deep models that cannot provide explanations out of the box.

We note that there is no confusion about interpretation, explainability and semantics in the case of the history of mathematical logic (e.g., \cite{mendelson2015}). When the vocabulary of the representation (well-formed formulae) is precise, interpretability is obtained by ensuring that each component is assigned a fixed interpretation (e.g., constants to individuals in a world, variables range over constants, truth values to logical connectives, etc.). And the semantic interpretation of {\em{any}} expression is determined compositionally by interpretation of an expression's components. 

But the manner in which representations emerge in the context of empirical developments in machine learning has not typically been guided by any adaption of extension of the systems of interpretability and semantics of logic. Our perspective is that the principles of mathematical logic can be easily adopted to a broad range of machine learned representations, in order to help humans understand learned representations.

In this context, an interpretable model is one that a human user can read or inspect, and analyze in terms of composable parts. In this way, interpretability refers to a static property of the model, and can vary from fully interpretable (models such as a small decision tree), to deep neural network models in which interpretability is more complex and typically limited. For instance, consider what each layer learns in a convolutional neural network (CNN): early layers are responsible for extracting low-level features such as edges and simple shapes, while later layers usually extract high-level features whose semantics are understood with respect to an application domain. In fact, with this perspective, models such as deep neural networks could hardly be classified as interpretable. It is important to point out that interpretability applies to the interpretation a learned model before considering the inference the model can do. Note that we are against classifying models as interpretable or non-interpretable, but rather we believe there should be a spectrum allowing an interpretability score to be assigned to each model.

\begin{figure}[h]
    \centering
	\includegraphics[width=0.95\textwidth]{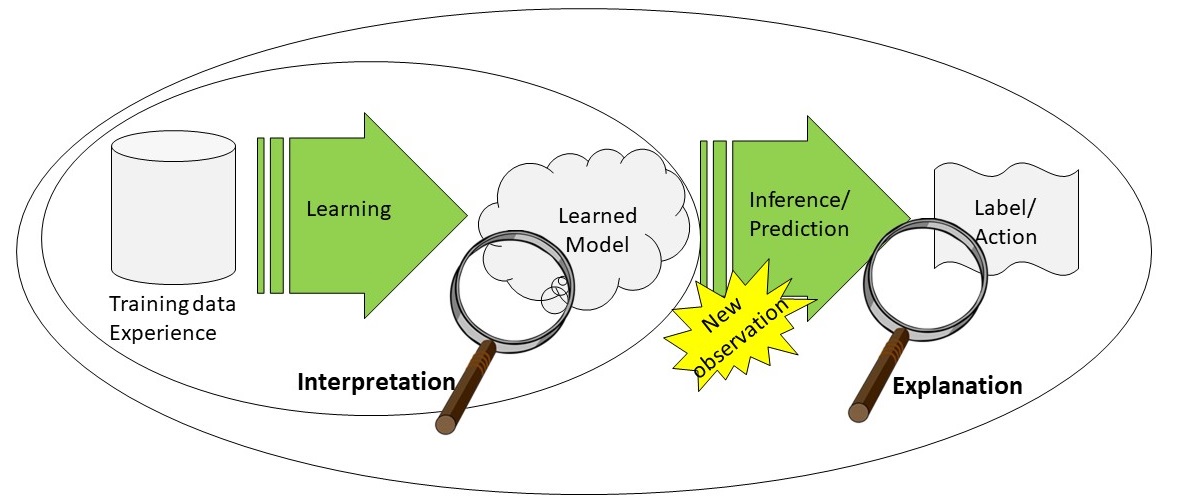}
	\caption{Interpretability of a model vs. Explainability of a prediction.}
	\label{fig:explain}
\end{figure}

On the other hand, explainability has to deal with what kind of output the system provides to the user, rather than how a human user directly interprets the meaning of each model component. In other words, explanation has to  do with clarifying the reason or reasons a prediction was made or an action was taken. Thus, we define an explainable model as a system which is capable of providing explanations without doing any extra computation. {\em Explainability} is, thus  a dynamic property of a model, in the sense that it requires runtime information to produce explanations. Explainability pertains to the mechanism of justification provided for an inference or prediction using a learned model, whether the used model is clearly interpretable or loosely interpretable. Figure \ref{fig:explain} illustrates the distinction between interpretability, which concerns the rendition or comprehension of a predictive model learned from data during training, and explainability, which pertains to the elucidation and justification of a prediction or decision made in the presence of a new observation or case. Both may revert to and rely on the original training data for analogy or grounds for justification.

Based on the above definitions, models such as decision trees and rule-based systems that are considered transparent (i.e., they are generally considered toward the fully-interpretable end of the transparency spectrum) are also explainable, while deep models are not. For example, once we add an explanation module to the deep neural model (e.g., \cite{babiker2017}), they become explainable systems as well but, interpretation of their meaning can be either in terms of how models are learned or what predictions mean (i.e., their interpretation in their application domain). 

Notice that these distinctions between explainability and interpretability do not comply with a reasonable assumption that is true in a common sense usage of the terms: outside of the AI arena, it is reasonable to expect explainability requires interpretability; one can only explain something they fully understand and can interpret. That stems from our definition of explainability: in AI, an explanation carries a completely different meaning from the one of its usual usage. Even a saliency map which highlights areas of an image is considered to be (to some extent) an explanation of an image classification system. That ``explanation'' does not consider image semantics, it is just an objective identification of which pixels contribute more to the final activations in a neural network. Still, they can be especially useful as debugging tools for machine learning practitioners (see Subsection~\ref{subsec:debug_vs_explanation}).

\subsection{Alternative explanations: who are explanations for?}
\label{subsec:explainee}

According to \cite{Lent2012}, a person's background knowledge, often called prior knowledge, is a collection of ``abstracted residue'' that has been formed from all of life's experiences, and is brought by everyone, at any age, to every subsequent life experience, as well as used to connect new information to old. 

As such, it becomes clear that, in the context of XAI, systems should be able to effectively take background knowledge into consideration in order to connect predictions and predictive models, and to shape explanations to the appropriate level of detail, i.e., adjusting explanations to conform to the knowledge of the corresponding {\it explainee}. However, the most common current approaches to explainability in AI systems attempt to provide information on a model's inner functioning without regard for the consumer of that information (see Subsection \ref{subsec:debug_vs_explanation}). 

To illustrate the importance of considering the explainee (and hence his/her background knowledge, expectations, etc.), consider an interview by Richard Feynman with the British Broadcasting Corporation (BBC) in 1983, in which he was asked why magnets with the same poles repel each other when placed close enough \cite{Feynman_why}. Feynman argues that, to properly explain that behaviour, he would need to consider the reporter's background on that matter, and any answer provided could unfold a new round of questions and explanations; and this process could continue indefinitely as new details are provided to explain the previous answer. The point is, of course, that an explanation's satisfaction with this iterative dialogue is the foundation of how XAI systems should be evaluated (cf. Subsections \ref{subsec:explainee}, and \ref{subsec:measuring}).

\subsection{Debugging versus explanation}
\label{subsec:debug_vs_explanation}

As mentioned above, some approaches to explainability provide information related to how the model works internally. However, not all information provided by those approaches can really be considered domain explanatory information. Of course, the information provided by, e.g., rule-based systems can be understood as a detailed explanation on how the system operates and could be applicable in scenarios where an end user needs to understand how the system generated a prediction. However, other approaches (especially those applicable to the so called opaque or black-box systems) are way less informative, and can be considered superficial ``hints'' rather than actual explanations, e.g., saliency maps on convolutional neural networks. This is really an observation about understanding internal components so as to debug those mechanisms, not as explanation. Although explanation-wise constrained, these approaches are still useful on helping to understand how a model behaves, especially if the consumer of the information has the necessary background. Hence, they may be considered debugging techniques for opaque AI systems rather than production of explanations based on a user's understanding of the semantics of an application domain. 

One example of a debugging tool to augment a model is the work of \cite{lecue_kgxai}, in which the authors used a ResNet \cite{rcnn} to recognize objects in a scene. Applying a saliency map to figure out what area in the image was contributing more to the final activations is not really helpful for a (lay) human consuming the model output to understand misclassifications, but it may help a researcher at design time to figure out alternatives to overcome the model limitations. In this case, the authors augmented the model by post-processing the final results using an external knowledge graph to add semantic context and modify the confidence score of the recognized objects.

An alternative, perhaps more foundational model, is presented by Evans and Greffenstette in \cite{evans2018}, who articulate an inductive logic programming (ILP) framework in which explanatory models are identified in the space of all possible inductive logic programs. This framework requires the development of a measure space for all such ILP instances, in which a gradient can be determined.  But the positive consequences of that technical maneuver is that an instance of an inductive logic program can be interpreted at the level of the semantics of an application domain, all the way down to instructions for a Turing machine. This framework does not resolve the challenge of what an appropriate level of explanation should be for a particular explainee; but it does provide a rich and mathematically elegant space in which to identify everything from descriptions of computation to arrive at a predictive model all the way to rule-based specifications at the level of an application domain.

\subsection{Is there a trade off between explanatory models and classification accuracy?}

Deep learning-based systems became prevalent in AI especially after its successful applications in image classification problems. Deep learning-based systems achieve impressive accuracy rates on  standard datasets (e.g., ImageNet \cite{imagenet_cvpr09}) without requiring much effort on designing and implementing handcrafted rules or feature extractors. In fact, by leveraging transfer learning techniques and well known architectures based on convolutional neural networks, a deep learning practitioner can quickly build an image classifier outperforming image classification methods which were state of the art before the ``deep learning revolution.''

\begin{figure}[h]
    \centering
	\includegraphics[width=0.70\textwidth]{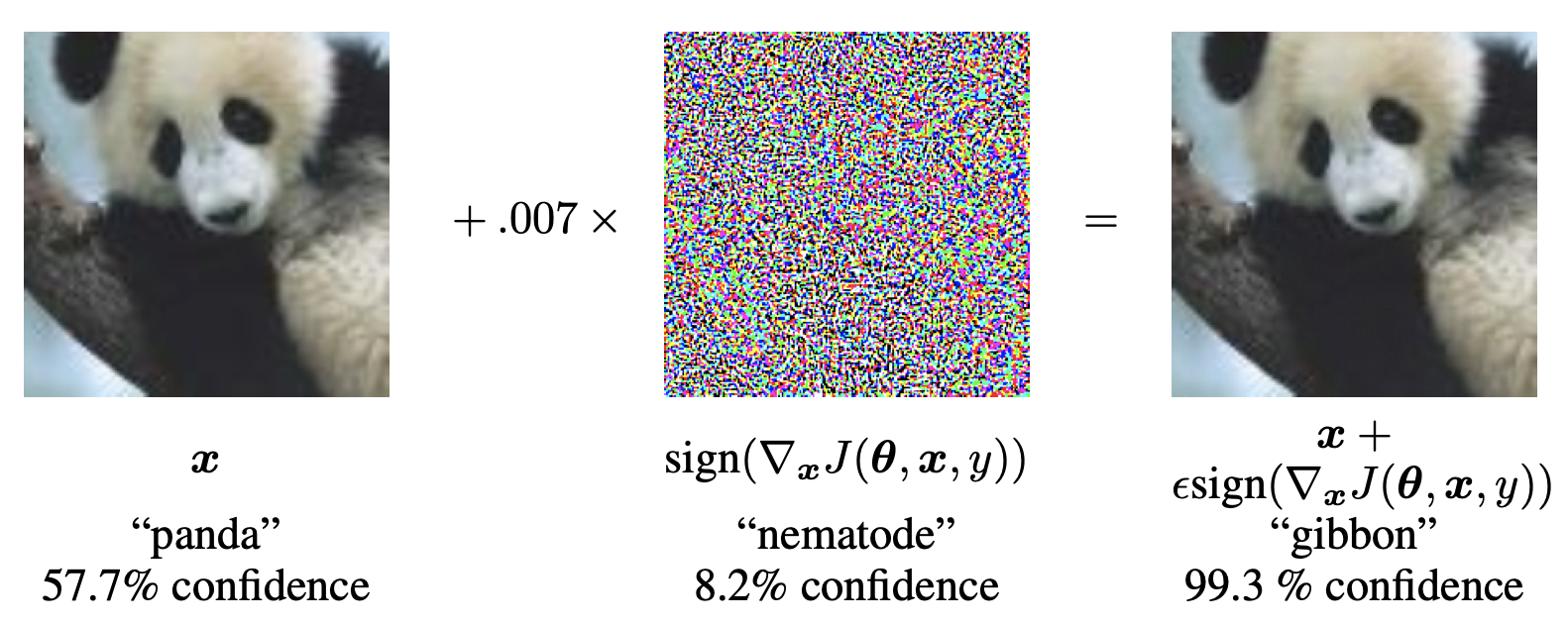}
	\caption{A panda image mistakenly classified as a gibbon after noise is added \cite{goodfellow2014explaining}.}
	\label{panda-gibbon}
\end{figure}

Nevertheless, despite their excellent overall accuracy, deep learning systems are considered black-boxes unable to provide explanations as to why they make a given prediction. In some applications, that limitation does not translate into a serious practical problem: a mobile phone picture classification application which misclassifies two animals will not bring consequences to a user other than a few giggles and a funny discussion topic at friends gatherings. If those errors are seldom, nobody would really care or lose confidence on the application. Errors may come up in random images and could be induced. Figure \ref{panda-gibbon} shows an example of a panda picture being classified as a gibbon after some adversarial noise is added \cite{goodfellow2014explaining}.

\begin{figure}[h]
    \centering
	\includegraphics[width=0.70\textwidth]{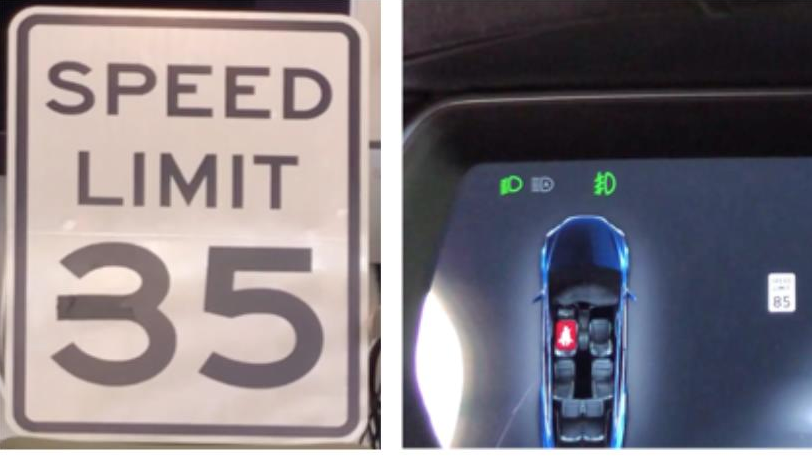}
	\caption{A modified speed limit sign reads as 85 mph on the Tesla's heads-up display \cite{teslaspeedlimit}.}
	\label{fig:teslaspeedlimit}
\end{figure}

The above example illustrates it is possible to intentionally fool a classifier through addition of appropriate noise. Depending on the image classification application, that kind of error may produce more serious consequences than the hypothetical phone application mentioned above. For example, recently hackers were able to fool Tesla's autopilot by tampering speed limit signs with adhesive tape (see Figure \ref{fig:teslaspeedlimit}), making the car to accelerate to 85 mph.

This is a simple example which illustrates predictions from AI models cannot be blindly accepted in many practical applications. Moreover, techniques unable to explain how they arrive at a prediction make them even more sensitive to random errors or deliberate attacks. That observation raises an important question around a potential trade off between model accuracy and explanatory capabilities: it is true that a deep learning-based model can achieve accuracy in many practical applications. That allows practitioners to quickly build accurate models with not so much effort. However, some preconditions do exist, the main one being the availability of potentially large labelled datasets (a problem potentially alleviated by transfer learning, but still common in machine learning in general and in deep learning techniques in particular). In some cases, training large state of the art deep learning networks requires thousands of even millions of dollars (the estimated cost of training just one of models developed in \cite{adiwardana2020humanlike} was estimated in US\$1.4 million \cite{meenacost}). All considered, it is not appropriate to claim there is necessarily a trade off between accuracy and explainability (or more generally, model performance). In some cases, deep learning methods will not be able to provide state of the art results (e.g., when there is not enough labelled data, when the model is so large it will be impractical to deploy on the target platforms, or even train due to prohibitive costs, etc.) so more explanation capable techniques might even provide better results. But as previously noted, there is no reason in principle that induced models like decision trees should in principle be less accurate than deep learned models.

\subsection{Assessing the quality of explanations}
\label{subsection:assessing quality}

Whereas a factually wrong explanation is obviously inappropriate, determining if an explanation is good transcends its correctness. The quality of an explanation is a little like beauty; it is in the eye of the beholder. It is very clear (and quite common) that two factually correct, but different explanations could be considered good or bad depending on to whom they were provided. This means that, to assess quality of explanations, one (again) needs to consider the explainee, the person who receives the explanation (see Subsection \ref{subsec:explainee}). The explainee's background, expectations, goals, context, etc., will play a determinant role in the evaluation process.

From the above paragraph, it is clear that assessing the quality of explanations is subjective, and a quite complicated task, even if done manually. Thus, coming up with an effective technique to evaluate explanation capabilities is beyond the reach of currently available methods. In fact, automatic evaluation of any generative model is a difficult task. Metrics commonly used for translation systems such as BLEU \cite{Papineni2002bleu} or for automatic summarization such as ROUGE \cite{lin2004rouge} are not appropriate for more sophisticated tasks such as explainability or even dialogue systems, since they assume that valid responses have significant word overlap with the ground truth responses \cite{liu2016_diagsys}. 

For that reason, most evaluation methods for explainability systems require human intervention. For example, the organizers of a fake news detection competition\footnote{https://leadersprize.truenorthwaterloo.com/en/} which requires an explanation of why a given statement is considered fake news or not, split the competition in two phases and limited the explanations assessment to the second phase to which only 10 teams would be qualified, thus making it manually tractable. 

The history of evaluation in the field of data visualization is also relevant to the question of how to evaluate explanations. The initial focus on alternative visual renderings of data have, over a decade, transformed from whether a visualization was ``interesting'' to consideration for what human inferences are enabled by alternative visualization techniques (e.g., \cite{Spence2014}).

The simplest conceptual alignment is that a visualization is a visual explanation.  The compression of a large volume of data to a visualization picture is lossy and inductive, so the choice of how to create that lossy inductive picture or explanation is about what inferences to imply for the human visual system.  The evaluation of alternative visualizations has evolved to a framework where evaluation is about what inferences are easily observed (e.g., \cite{Lam2012}).  Furthermore, interactive visual explanation is easily considered as our suggestion of explanation being interactive and driven by the semantics of the application domain (e.g., \cite{Goebel2013}).

Evaluation of what we can consider as a more general explanatory framework, which produces alternative explanations in terms of text, rules, pictures, and various media, can similarly be aligned with the evolution of how to evaluate visual explanations.

But of course there is yet no clear explanation evaluation framework, but only a broad scope of important components (e.g., \cite{miller-explanation-2019}).  Even specific instances of proposals for explanation evaluation beg the need for increased precision.  For example, \cite{adiwardana2020humanlike} suggest explanation quality is dependent on two main factors: sensibleness and specificity. A measure which takes those factors into account (Sensibleness and Specificity Average - SSA). This suggestion arose from work on the topic of dialogue systems, and has been characterized in terms of a high correlation with another measure called ``perplexity:'' a measurement of how well a probability distribution or probability model predicts a sample. A low perplexity indicates the probability distribution is good at predicting the sample. In the context of conversational systems, perplexity measures the uncertainty of a language model, which is a probability distribution over entire sentences or texts. The lower the perplexity, the more confident the model is in generating the next token (character, subword, or word). Thus, perplexity can be understood as a representation of the number of choices the model is trying to choose from when producing the next token. This measure is commonly used to assess the quality of conversational agents and as a metric which must be optimized by machine learning based dialogue models. Thus, although not ideal and lacking specific experiments on the domain of explainability, perplexity could potentially be effectively used to evaluate text-based XAI systems as a reasonable approximation of human evaluation.

While we have more to say about evaluation below, what is clear is that evaluation of explanatory systems is based on how the explainee confirms their own understanding of an explanation or the conclusion of an explanatory dialogue.

\section{A brief history of explanation}
\label{section:history}

\subsection{Abduction}
Explanations have always been an indispensable component of decision making, learning, understanding, and communication in the human-in-the-loop environments. After the emergence and rapid growth of artificial intelligence as a science in the 1950s, an interest in interpreting underlying decisions of intelligent systems also proliferated. Especially, C.S. Peirce’s hypothesis of abduction \cite{peirce1891architecture} stimulated the AI community’s attention to exploiting this conceptual framework for the design and development of complex expert systems in a variety of domains. Abduction or abductive reasoning is a form of reasoning that starts with a set of observations and then uses them to find the most likely explanations for the observations.

A compressed historical journey of Peirce's ideas can be traced in four projects, beginning with Pople \cite{Pople1973}, Poole et al. \cite{poole1987}, Muggleton \cite{muggleton1991}, to Evans et al. \cite{evans2018}. In 1973, Pople provided a description of an algorithm to implement abductive and showed its application to medical diagnosis.  Poole et al. extended abductive ideas to a full first order implementation and showed its application to guide the creation of explanatory hypothesis for any application domain.  Muggleton produced a further refined system called inductive logic programming, in which creation of hypotheses are generally identified by inductive constraints in any general logic.  Finally, the adoption of this thread of mechanisms based on abductive reasoning have been generalized to the full scope of explanation generation based on inductive logic programming by Evans et al.   Every instance of these contributions relies on a logical architecture in which explanations arise as rational connections between hypotheses and observations (cf. scientific explanation).  The most recent work by Evans et al. extends the framework in a manner that supports modern heuristics of inductive model construction -- or learning of predictive models -- by providing the definition of a gradient measure to guide search over alternative inductive logic programs.

In fact, that thread of exploiting abduction in Artificial Intelligence is aligned with perspectives from other disciplines. For example, Eriksson and Lindstr{\"o}m describe abductive reasoning as an initial step of inquiry to develop hypotheses where the corresponding outcomes are explained logically through deductive reasoning and experimentally through inductive reasoning \cite{eriksson1997abduction}.  Their application to ``care science'' is just another example that confirms the generality of abductive reasoning.

\begin{figure}[H]
    \centering
	\includegraphics[width=0.70\textwidth]{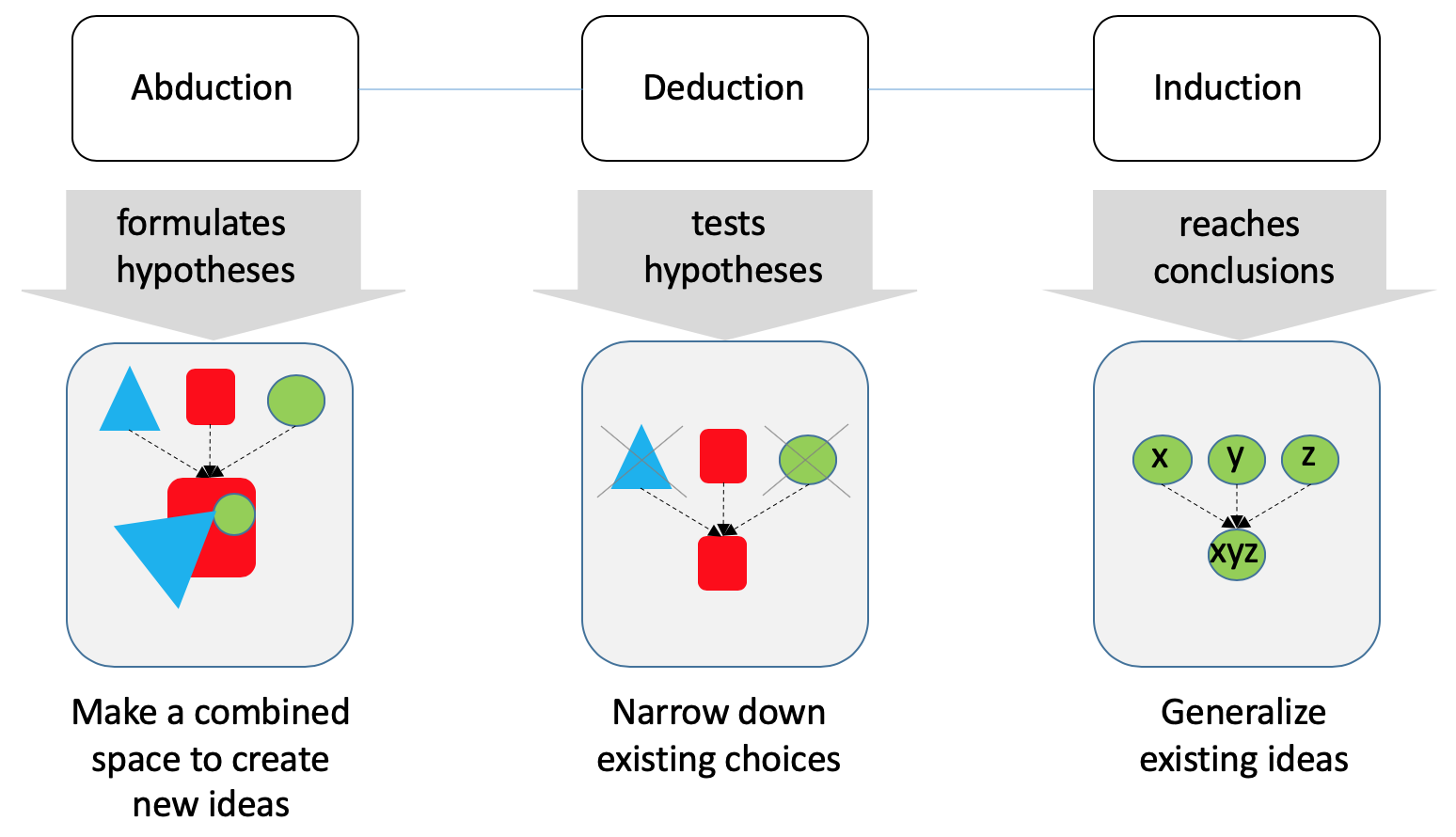}
	\caption{The process steps of the reasoning methods}
	\label{fig: abduction components}
\end{figure}

The block diagram of Figure~\ref{fig: abduction components}, partially inspired by a figure in \cite{abductionfigure}, is intended only to confirm the connection between abductive, deductive, and inductive reasoning.
We see that abductive reasoning entails justification of ideas that support the articulation of new knowledge by integrating deductive and inductive reasoning. In Artificial Intelligence studies, the process involving these reasoning steps are as follows: 1) identify observations that require explanation as they cannot be confirmed with already accepted hypotheses; 2) identify a new covering hypothesis using abductive reasoning; 3) empirical consequences of the hypothesis, including consistency with already known knowledge, is established through deduction; 4) after an accepted level  of verification, the hypothesis is accepted as the most {\em scientifically plausible}.

\subsection{Scientific explanation}  \par

The connection between Artificial Intelligence frameworks for abductive explanation have suggested a direct connection between  ``scientific explanation,'' and is the subject of many debatable issues in the community of science and philosophy \cite{woodward2003scientific}. Some of the discussions imply that there is only one form of explanation that may be considered scientific. There are also some proponents of the idea that a theory of explanation should include both scientific and other simpler forms of explanation. Consequently, it has been a common goal to formulate principles that can confirm an explanation a scientific explanation. As far back in history as Aristotle, generally is considered to be the first philosopher to articulate an opinion that knowledge becomes scientific when it tries to explain the causes of ``why''. His view urges that science should not only keep facts, but also describe them in an appropriate explanatory framework \cite{falcon2006aristotle}.

In addition to this theoretical view, empiricists also maintain a belief that the components of ideas should be acquired from perceptions with which humans become familiar through sensory experience. The development of the principles of scientific explanation from this perspective prospered with the so-called Deductive-Nomological (DN) model that was described by Hempel in \cite{hempel1942function}, \cite{Hempel1958-HEMTTD}, \cite{hempel1965aspects}, and by Hempel and Oppenheim in \cite{hempel1948studies}.

The DN model is based on the idea that two main elements form a scientific explanation: An \textit{explanandum}, a sentence that outlines a phenomenon to be explained, and an \textit{explanan}, a sentence that is specified as explanations of that phenomenon. For instance, one might constitute an explanandum by asking ``Why did the dish washer stop working?'' and another person may provide an explanan by answering ``Because the electricity went off.'' We may infer that the explanan is rationally or even causally connected to the explanandum, or at least that the explanandum is the reasonable consequence of the explanans, otherwise speaking \cite{mcgrew2009philosophy}. In this way, the explanation delivered as an explanans becomes a form of deductive argument and constitutes the ``deductive'' part of the model. Note that a series of statements comprises an explanan should comply with ``laws of nature.'' This is a vital property, because derivation of the explanandum from the explanan loses its validity if this property is violated \cite{woodward2003scientific}. This is the nomological component of the model, where the term ``nomological'' means ``lawful.'' Hempel and Oppenheim’s DN model formulation states that a scientific explanation is an answer to a so-called ``why'' question, and there may be multiple such answers.   There may also be several types of questions (e.g.,  ``How does an airplane fly?'') that cannot be converted into why-questions. Consequently, answers to such questions are not considered to be scientific explanations. This does not mean that such answers are not part of a scientific discipline; these answers just become \textit{descriptive} rather than being \textit{explanatory} \cite{bunzl1993context}, and is related to our next section on causality.

Another aspect of the DN model is that the elements of an explanation are statements or sentences describing phenomenon, not the phenomenon itself. Finally, the sentences in the explanans must be accurate and verified, urging the arguments of the scientific explanation to be \textit{valid} and \textit{sound}. Thus, the DN model can be summarized as a model of a scientific explanation outlining a \textit{conception} of explanation and a \textit{connection} in the flow of an explanation.

\subsection{Causality} 

As hinted in the summary of scientific explanation, perhaps the most strict form  of explanation is causal explanation.  Informally, a casual explanation is one that arises from the construction of causal models, which require that explanations for arising predictions are in face ``recipes'' for reconstruction that prediction.

Causal models typically facilitate the creation of explanation for a phenomena or an answer to a query by constructing a formal expression. That formal expression is derived from some causal representation, which typically captures directed causal relationships in a graphical model of cause and effect (or causality). This representation encodes an incomplete set of assumptions built upon prior knowledge. The causal explanation expression is, as with abduction and scientific explanation,  revised continuously until a suitable explanation is obtained,  and can answer the particular query. 

The most relevant and recent framework of causal representation and reasoning is given by the culmination of Pearl's research in \cite{pearl2018book}.  In that work, an abductive explanation is called an ``estimand.'' This idea of a formal expression that best explains a particular phenomena (or a query) has its root in formal philosophy, as noted about, and especially in abductive reasoning. As noted above, abductive reasoning is given a set of incomplete observations (or assumptions as described above) and seeks to construct an explanation which best describes it (or the estimand).

An important point here is that the overall information architectures of abduction, scientific reasoning, and causal reasoning are similar, but their mechanism and the evaluation of an explanation are successfully refined.

\subsection{Explaining mechanism/syntax versus semantics } 

A lingering unaddressed distinction is about the content or meaning of an explanation, especially in the context of what counts as an explanation to a user.  Again a principled distinction exists in the realm of mathematical logic (cf. \cite{mendelson2015}; any logic textbook will suffice). In the context of predictions from domain models (whether learned or fabricated by hand), a prediction has at least two kinds of explanation.  For example, consider the simple familiar syllogism

\begin{itemize}
\item All men are mortal.
\item Socrates is a man.
\item Socrates is mortal.
\end{itemize}

Consider ``Socrates is mortal'' as a prediction of the very simple model. From the perspective of formal logic, there are (at least) two explanations. One is the explanation of the deductive mechanism that produced ``Socrates is moral'' from the first two expressions. This a so-called proof-theoretic explanation as it amounts to a description of how two premises are combined by deductive inference to derive the prediction.
In an analogy with programming language debuggers, this kind of explanation is about the mechanism that produced the prediction, and is akin to how current work in explaining image classification (e.g., \cite{babiker2017}).  This kind of explanation is appropriate when the explainee has interest in understanding and debugging the mechanism.

But note an alternative explanation is not about mechanism but about the meaning of the expressions. Logically, the proof theory or deductive chain explanation is about mechanism.  But the semantic explanation is about what it means to be mortal and what it means to be a man.  That kind of explanation is semantic, and is intended to be appropriate for an explainee who is not interested in mechanism but in meaning.  If a prediction was ``Socrates is a duck'' obtained from the same system, it can immediately be viewed with suspicion because of its meaning, not because of the mechanism that produced it from a presumably faulty model.

So distinguishing syntax from semantics or meaning has more to do with the internal rules that a system has to follow to compute something. We all know that symbolic debuggers for programming languages create labels and traces which become the basis for producing mechanism explanations. The computation rules themselves might not be sufficient to provide a clear picture on why a system came to a conclusion (or an answer to a query). But interpretation of syntactic expressions is what creates asemantic interpretation. Returning to the idea of estimands, an estimand can be viewed as a well-constructed expression if it makes sense semantically.
As with the simple syllogism above, the form of the explanation can be based on that of the causal (or any) model.

In this era of deep learned models, we can consider these relationships between syntax and semantics as the internal representations of each layer and their composition at the final layer respectively. Interestingly, this notion of construction of semantics (whole) as a function of semantics of its parts and their careful combination that obeys a particular syntax is very familiar in the logics to interpret natural language, developed by a famous linguist-logician, Richard Montague \cite{stanford-encyclo-phil}. At the syntactic level we might infer the correlation among different variables (in the intermediate layers) of a deep learned system but in semantic level we know what combination of those variables (in the final layer) provide an interpretation for a particular query. Ontology driven explanation for a ResNet model \cite{lecue_kgxai} described in Section~\ref{section:principlecomponents} is one good example of the use of semantics to explain an opaque system.

\section{Classification of Current Research Trends}
\label{section:trends}

In the last five years, there has been a surge in the papers attempting to introduce new explanation methods. This intensity of work in XAI is, in fact, a side effect of widespread use of AI in sensitive domains such as legal reasoning and the medical field. In this section, we review some of the various explanation approaches popular in the literature, and classify in our framework based on how the explanations are built, and compare that with the levels of explanation introduced in Section \ref{section:intro}. 

\subsection{Concurrently constructed explanations}
Some have focused on creating models that try to build explanations concurrently together with the  main task (e.g., learning a classifier). As an example, consider the work of \cite{lei-etal-2016-rationalizing} who seek to identify segments of text that support an explanation of text review classification. 
Their approach proposes a neural architecture that is made up of a generator followed by an encoder component. The generator extracts portions of the input text as salient words, then forwards them to the encoder to predict the target class. The final output of the system comprises the class label, together with the extracted ``justification'' from the input text. Other similar work, applied beyond text to images and text, has relied on learning attention weights from the input. In the related work, some authors referred to this category as learning interpretable representations.  

In natural language processing (NLP) text classifications for instance, attention layers attempt to learn the weight of each latent representation produced by the recurrent layer. The attention weights are then used to explain the prediction made by the classifier \cite{yang2016hierarchical}. There is a debate in the literature on whether attention weights could be used as an explanation or not \cite{serrano2019attention, jain2019attention}. An interesting connection is to our discussion above regarding the difference between debugging explanations and semantic explanations; much of this research is motivated to equate a mechanism behaviour to semantic interpretability.

\subsection{Post-hoc explanations}
Another approach is to use a post-hoc technique.  The basic idea is to approximate explanations from  a trained model. 
As mentioned earlier, concurrently constructed explanations need to be computed within the model, which means they need to have access to the internals of the model, or what many refer to as  model-dependent (this further creates confusion about whether a model is syntactic or semantic). However, some post-hoc approaches can create approximate explanations without having access to the internals of the model, thus could be classified either as model-dependent or model-independent\footnote{Sometimes, this is called model-agnostic post-hoc explanations.}. In the next subsection, we will briefly discuss the difference between model-dependent and model-independent. 
\subsubsection{Model-dependent explanations}
To describe model-dependent explanations, consider the case of non-linear deep networks. One can use a back-propagation algorithm to learn feature importance (e.g., which pixels contributed most in classifying the image as a cat rather than a dog) then use that learned feature ranking as the basis for explaining predictions.
The simplest general approach is to compute a gradient with respect to the predicted class and use the back-propagation to propagate the gradient to the input. 
Finally, one can combine the input with the gradient to capture the salient pixels which can be used to explain the predicted class (e.g., Grad-CAM \cite{selvaraju2017grad}). 

\subsubsection{Model-independent explanations}
The goal of this group of methods is to focus more on explaining individual instances without the target model being exposed. In fact, the target model is now a black-box model. Ribeiro et al. introduced LIME \cite{ribeiro2016should} to approach the explanation problem using a perturbation method. They perturb the original data point to create a new dataset in the vicinity of that instance. The black-box model is queried to get the labels associated with the aforementioned points. This labelled dataset is then used to frame a near enough justification. While LIME is the most cited model-independent method, there are other approaches which can be classified as model-independent \cite{lundberg2017unified,guidotti2018local}.

\subsection{Application-dependent vs. generic explanations}
\label{subsection:application_dependent}
Another way to classify explanation methods is to consider how an explanation mechanism is related to the application domain of the task. An application-dependent method implicitly assumes the explainee is knowledgeable about the application and thus it employs the domain's vocabulary. In a medical application, for instance, a system can explain the prediction using medical terms. A generic explanation, on the other hand, can only provide explanations based on the mechanism of model building, combined with information available in the training set (e.g., correlation between features). Note that a model-dependent method is not necessarily taking into account the knowledge of the explainee (i.e., it will provide the same explanation irrespective of the customers' knowledge), but it must take advantage of the application's vocabulary (see Subsection~\ref{subsec:debug_vs_explanation}). It is also noteworthy that the system needs to go beyond correlative features --- which is how most current machine learning methods work --- to be capable of providing such application-dependent explanations. Many explainees (e.g., physicians, lawyers) would prefer having application-dependent explanations. This will not be achieved without moving the machine learning research on explanation toward scientific and casual explanation.

\subsection{Classification based on levels of explanation}
As described briefly in Section 1, different levels of explanation could be introduced as shown in Figure~\ref{fig:explainability-diagram}. Here we want to further elaborate  those abstract levels and classify the related work accordingly. Table~\ref{tab: classification_of_research_threads} classifies some of the most prominent existing work based on the levels of explanation. Most of recent research have focused on Level~1, and only a few have worked on Level~2. To the best of our knowledge there is as yet no existing work on Levels~3 and 4. However, there are some conceptual approaches that aim to achieve such levels \cite{madumal2018towards}.  In the subsections below, we provide details.
\begin{table}[h!]
    \begin{center}
        \resizebox{0.9\textwidth}{!}{\begin{minipage}{\textwidth}
        \begin{tabular} {l|ccccc}
        \hline
            \textbf{Method}  & \textbf{Level} & \textbf{MD/MI} & \textbf{CC/PH}\\
            \hline
            LIME \cite{ribeiro2016should}     & 1 & MI & PH \\
            Grad-CAM \cite{selvaraju2017grad}     & 1 & MD  & PH  \\
            SHAP \cite{lundberg2017unified}     & 1 & MI & PH  \\
            Rationalizing-predictions \cite{lei-etal-2016-rationalizing}     & 1 & MD & CC \\
        \hline
        Grounding visual explanations \cite{hendricks2018grounding}     & 2 &  MD & PH   \\
        \hline
        \end{tabular}
        \end{minipage}}
    \end{center}
    \caption{Classification of recent explanation techniques based on Levels of Explanation. MD stands for Model-Dependent while MI means Model-Independent. CC corresponds to Concurrently Constructed explanations and PH refers to Post-Hoc technique.}
    \label{tab: classification_of_research_threads}
\end{table}
\subsubsection{Level 0}
Models classified as Level~0, provide no explanation at all. They are, in essence, black-box models that cannot provide any explanatory information to a user. In other words, the explainee is expected to accept or reject a system's decision without any further information. Most  off-the-shelf methods for learning classifiers (e.g., deep learned models, support vector machines, or random forests) belong to this level. 

\subsubsection{Level 1}
The explainee is provided with a single type of explanation in models falling into this category. For example, a framework that provides heat-maps to explain image classification belongs to this level. Most of this approach focuses on providing a post-hoc explanation, which transitions a black-box model ---that originally belonged to Level~0--- to a Level~1 model. Recently, however, a few methods have been proposed to look at building concurrently constructed explanation algorithms \cite{lei-etal-2016-rationalizing, bastings2019interpretable} to make models that by definition belong to Level~1.

\subsubsection{Level 2}
Level 2 adds another type of explanation to enrich the knowledge communicated with the explainees. At this level, the system not only provides a heat-map to explain a classified animal image as a cat, but it also contains another type of explanation such as a textual explanation as an alternative description of the predicted classification. In this way, the alternative explanations allows the user to grasp more insights about the reasoning process employed by the system to make the prediction. If one explanation is not well understood by the explainee, then they have the opportunity to understand from an alternative explanation. Note that in the case of the abductive systems described above, there can be a large number of alternative explanations.

\subsubsection{Level 3}
An explainee and their familiarity with the domain plays a vital role in this level. The explanatory system includes some model of the explainee's domain model, and is capable of deciding the right type of explanation according to the knowledge of the explainee. For instance, a patient is diagnosed with some disease and an AI system is used to provide a potential treatment therapy. While the therapist requires a detailed medical explanation by the AI system, the patient would strongly prefer to have a lay person's explanation for any alternative treatment recommendations.

In the current context of the COVID-19 pandemic as an example, Hydroxychloroquine is alleged to be a potential cure and has attracted many ordinary people's attention around the globe. People are interested to understand why this drug is a potential treatment. As a result, many medical researchers provide interviews to the media explaining how this drug works, typically with very shallow detail. As we can expect, however, the same experts would use a different level of granularity to explain the drug to other experts.
Please note that, none of the existing explanation methods take into account the knowledge of the explainee.

\subsubsection{Level 4}
While previous levels (e.g., Level~0, 1, and 2) do not include the capability of interaction between the explainee and the system except perhaps for at most one interaction (Level~3), methods classified as Level~4 can interact with the user. They are expected to support a conversation sort of capability which allows the explainee to refine their questions and concerns regarding the decision. In other words, each interaction in the conversation allows the explainee to get clarifications. Here, the system is capable of adapting its explanation to the vocabulary of the explainee. Take the Richard Feynman's interview \cite{Feynman_why} with the BBC as an example. He could provide the reporter with what he thought the reporter would understand most. Once the reporter understood that explanation, if he had further questions, or wanted more in-depth explanation, the reporter could ask, and an appropriate explanation could be provided by Richard Feynman. To the best of our knowledge, existing systems lack this interaction capability.

\section{Priority Components for a synthesis of an XAI architecture}
\label{section:components}

\subsection{XAI architecture}

As noted, much of the work on the explainability has focused on deep supervised learning, which describe  methods that answer the following two questions: (1) which input features are used to create an output prediction, and (2) which input features are semantically correlated with the outcome prediction.  The answers for these two questions contribute to the trust in the system, but explanation additionally requires a social process of transferring knowledge to the explainee considering the background knowledge of the explainee.

While the answers to questions (1) and (2) may acknowledge the importance of features that a model uses to arrive at a prediction, it may not necessarily align with a human explanation; prior knowledge, experience and other forms of approximate reasoning (e.g., metaphorical inference) may further shape an explanation,  while the predictions of a machine learning model may be restricted to the dataset and the semantics around it. Generally, an explanation system (for example, a human) is not restricted to the knowledge on which they make predictions and explanations and can draw parallels with different events, semantics and knowledge.

So merely responding to questions (1) and (2) do not satisfy the multiple purposes that XAI researchers aim to achieve: to increase societal acceptance of algorithmic decision outcomes, to generate human-level transparency about why a decision outcome is achieved, and to have a fruitful conversation among different stakeholders concerning the justification of using these algorithms for decision-making \cite{Kasirzadeh2019mathematical}.

To incorporate an interactive ``explainer'' in  XAI, an emerging XAI architecture needs to embed both an explainable model and an explanation interface. The explainable model includes all types of the pre-hoc, post-hoc and concurrent explanation models. As  examples of the explainable model, there can be a causal model, an explainable deep adaptive program, an explainable reinforcement learning model, etc. An explanation interface can be also a variety of types, such as a visualization system, or a dialogue manager with a query manager and a natural language generator that corresponds to Level-3 and Level-4 of Figure~\ref{fig:explainability-diagram}.    

\subsection{User-guided explanation}

As Miller \cite{miller-explanation-2019} notes, the process of explanation involves two processes: (a) a cognitive process, namely the process of determining an explanation for a given event, called, as with Hempel, the explanandum. This identifies causes for the event, and a subset of these causes are selected as the explanation (or explanans); and (b) a social process of transferring knowledge between explainer and explainee, generally an interaction between a group of people, in which the goal is that the explainee has enough information to understand the causes of the event. This is one kind of blueprint for the Level~4 interactive explanation process noted above.

Miller  provided an in-depth survey on explanation research in philosophy, psychology, and cognitive science. He noted that the latter could be a valuable resource for the progress of the field of XAI, and highlighted three major findings: (i) Explanations are contrastive: people do not ask why event $E$ happened, but rather why event $E$ happened instead of some other event $F$; (ii) Explanations are selective and focus on one or two possible causes and not all causes for the recommendation; and
(iii) Explanations are social conversation and interaction for transfer of knowledge, implying that the explainer must be able to leverage the mental model of the explainee while engaging in the explanation process. He asserted that it is imperative to take into account these three points if the goal is to build a useful XAI.

One should note that it is plausible, given the study of explanation based on cognitive norms, that an explanation may not be required to be factual, but rather only to be judged to be satisfactory to the explainee (cf. Subsections~\ref{subsection:assessing quality}, and ~\ref{subsection:application_dependent}).

As we described in Figure~\ref{fig:explainability-diagram}, a dialogue system that can process a question of ``what if another condition'' from an explainee and produce a new prediction output based on the new condition will achieve another higher level of explanation. The explanation that can deal with ``What would the outcome be if the data looked like this instead?'' or ``How could I alter the data to get outcome X?'' is called contrastive explanation. Contrastive explanation is a human-friendly explanation as it mimics human explanations that are contrastive, selective, and social. 

To accommodate the communication aspects of explanations, several dialogue models have been proposed.  Bex and Walton \cite{Bex2016combining} introduce a dialogue system for argumentation and explanation that consists of a communication language that defines the speech acts and protocols that allow transitions in the dialogue. This allows the explainee to challenge and interrogate the given explanations to gain further understanding. Madumal et al. \cite{madumal2018towards} also proposed a grounded, data-driven approach for explanation interaction protocol between explainer and explainee. 

\subsection{Measuring value of explanations}
\label{subsec:measuring}

The production of explanations about decisions made by AI systems is not the end of the AI explainability debate. The practical value of these explanations, partly, depends on the audience who consumes them: an explanation must result in an appropriate level of understanding for the receivers of explanations. 
In other words, explanations are required to be interpreted and judged against different points, about whether they are good or bad, satisfactory or unsatisfactory, effective or ineffective, acceptable or unacceptable.

Again the previously mentioned evolution of the evaluation of visualization systems is highly relevant, as that evolution ultimately requires the design of cognitive experiments to confirm the quality and value of alternative explanations, visual or not (see Subsection~\ref{subsection:assessing quality}).  It is clearly the case that quality of a ``visual explanation'' is about how well it leads the reader to the intended inferences from the visualized data domain. Naturally, the background knowledge of a viewer is like the background knowledge of an explainee; their knowledge and experience determines what preferred inferences obtain.

Looking forward to how to evaluate XAI systems, among those background assumptions that impact the judgements of explanations are what are returned to as cognitive ``norms.'' It has been empirically shown that norms influence causal judgements \cite{adam2017norm}. To put it simply, norms are informal rules that are held by people, and can have statistical or prescriptive content.
The empirical and mathematical aspects for why a decision outcome is achieved are interpreted against some background assumptions held by the audiences of explanations. Some disagreements with an explanation for a decision outcome in a sensitive context due to the background assumptions of the audience of explanations reveal some moral or social mismatch about algorithmic decision-making between the receiver of an explanation and its producer. 

If one does not have an appropriate level of knowledge about the relevant precedent assumptions, one might not have the capacity to judge and interpret an explanation of a decision.  In that case, iteratively refined question-answer dialogue (cf. Fenyman's point made in Section~\ref{subsec:explainee}) may lead to an improved understanding by the explainee. In general the interpretability of explanations has significant practical value for revealing the explicit and the implicit reasons about why a decision-making procedure and process is chosen.

A schema for the interpretability of explanations aims to capture various precedent assumptions that become relevant in context-dependent evaluation of each kind of AI explanation for why a decision outcome is achieved.


Finally, in another elaboration of how to evaluate explanations \cite{DARPA}, there are proposed five measures of explanation effectiveness: (1) User satisfaction, (2) Mental model, (3) Task performance, (4) Trust assessment, and (5) Correctability.

User satisfaction is measured in terms of clarity of the explanation, and utility of the explanation. Task performance is to check if the explanation improved user's decision and task performance. Trust assessment is to assess trust and measure if it can be appropriate for future use. Assessment of a mental model is related to strength/weakness assessment, and it also assesses the predictions of ``what will it do'' or what if questions, and ``how do I intervene'' to adjust or guide explanatory outputs. Finally, Correctability is to measure if interaction with the system helps to identify and correct errors.  As far as we are aware, there is also no Level~4 system that has confirmed any experiments that demonstrate this kind of richly faceted evaluation.

Finally, to measure contrastive explanation that is close to human explanation, we need additional evaluation metrics for contrast, selection, and social explanation. Contrast can be measured in terms of the clear justification of the output through the comparison. Contrastive explanation should be able to explain why the output has been produced between the probable output candidates. Selection can be measured in terms of the importance (salience) of the reasons (features) that were mentioned during the contrastive explanation. Lastly, social explanation can be measured in terms of the clarity, understandability and utility of the explanation to the explainee. The measure of the social explanation corresponds to the measure of user satisfaction in \cite{DARPA}. But as noted, we know of no existing explanation systems that have been so considered with this rich palette of evaluation parameters.

\section{Summary and Conclusions}
\label{section:conclusions}

In summary, our goal has been to articulate a set of required components of an XAI architecture, and describe a high level framework to understand their connections.  In two alternative graphical depictions (Figure~ \ref{fig:explainability-diagram}, Table~\ref{tab:tabular_level_explainability}), we distinguish what we believe are mostly orthogonal components of an explanation system, and suggest an information framework related to levels of autonomous driving, where a richer set of components provides a more sophisticated explanation system.

That framework is descriptive and informal, but it allows us to factor some components (e.g., interpretability, explanation quality) into separate analyses, which we hope creates some line of sight to historical work on explanation.  No where is this more important than the history of abductive reasoning and its connection to the history of scientific reasoning, culminating in the construction and use of causal models as a basis for causal explanations.

We then try and consider more recent research in the context of these components and their relationship to the analysis of a deeper background literature, and provide some description of how those early ideas fit, and what they lack.  This culminates with a considering of how to evaluate explanatory systems, and connects recent work that addresses the cognitive properties of explanations.   Overall, we hope that our framework and analysis provides some connective tissue between historical threads of explanation mechanisms and modern reinterpretation of those mechanisms in the context of cognitive evaluation.

We conclude that there is much still to do to inform a principled design of a high level explanation system, but that there are many components
and integrating them with the appropriate knowledge representations within machine learning models, and respecting the cognitive aspects of evaluation, are a minimal requirement for progress.

\section*{Acknowledgements}

We acknowledge support from the Alberta Machine Intelligence Institute (AMII), from the Computing Science Department of the University of Alberta, and the Natural Sciences and Engineering Research Council of Canada (NSERC).

\bibliography{aij.xai}

\end{document}